# FiberGeoText: A Vision-Language Model for Population-Level Organization of Superficial White Matter

Yuqian Chen[1,2], R. Jarrett Rushmore[3], Guikun Chen[1,2], Fan Zhang[1,2], Edward Yeterian[4], Nikos Makris[1,2], Yogesh Rathi[1,2,5], Lauren J. O'Donnell[1,2]

1 Department of Radiology, Mass General Brigham, Boston MA, USA

2 Harvard Medical School, Boston MA, USA

3 Department of Anatomy and Neurobiology, Boston University Chobanian & Avedisian School of Medicine, Boston MA, USA

4 Department of Psychology, Colby College, Waterville ME, USA

5 Department of Psychiatry, Mass General Brigham, Boston MA, USA

**Corresponding Author**: Lauren J. O'Donnell (odonnell@bwh.harvard.edu) ; Fan Zhang (zhangfanmark@gmail.com)

**Acknowledgments**

This work was supported by the National Institutes of Health (NIH) grants: R01MH125860, R01MH132610, RF1MH132610, R01MH116173, R01NS125307, R01NS125781, R21NS136960.

**Conflicts of Interest**

The authors declare no conflict of interest.

**Data Availability**

The high-resolution gSlider data are our in-house dataset and will be publicly released via the NIMH data archive as part of a project funded by the National Institutes of Health (R01MH125860).

**Ethics Statement**

This study was approved by the institutional review board of Brigham and Women's Hospital. Informed consent was obtained from all participants.

# Abstract

The superficial white matter (SWM) is an important component of brain structural connectivity, with its significance highlighted by its roles in cognition across lifespan and brain disease. SWM contains abundant short-range association fibers whose organization remains incompletely characterized, in part because their short trajectories and close relationship to highly variable cortical folding make correspondence across individuals challenging. Anatomically corresponding connections may vary in spatial location across individuals and therefore may not be adequately defined by geometric proximity alone. We introduce FiberGeoText (FGT), a vision-language model (VLM) for organizing short-range superficial white matter (SWM) streamlines reconstructed from ultra-high-resolution diffusion MRI into population-level clusters. FGT jointly represents three complementary properties of each streamline: its three-dimensional trajectory, its cortical anatomical context, and its shape. Cortical endpoint information from multiple parcellation schemes is expressed as text and encoded using a pretrained large language model (LLM), enabling heterogeneous anatomical descriptions to contribute to a common continuous representation. These anatomical representations are integrated with learned streamline geometry and explicit length and U-ratio features to form a joint multimodal representation, which is subsequently used to derive a population-level SWM cluster organization. We evaluated FGT on acquired submillimeter 0.76 mm diffusion MRI data. Compared with state-of-the-art (SOTA) methods, FGT produced substantially greater cortical parcel coherence, within-cluster shape consistency, cluster-size consistency, and cross-subject correspondence, while allowing greater spatial variability than approaches optimized primarily for geometric compactness. Ablation experiments demonstrated complementary roles of the three information sources: cortical anatomy increased endpoint coherence, fiber geometry improved cluster compactness, and explicit shape features increased morphological consistency clusters. When applied to unseen subjects without reclustering, an average of 96.7% of the 5,000 learned clusters were recovered, and consistent cluster structure between the training and testing data was shown by embedding distortion ratios close to 1. Together, these findings demonstrate that population-level SWM correspondence can be learned by integrating complementary geometric, anatomical, and shape information, and demonstrate the value of multimodal deep embeddings learned from a VLM model for identifying population-consistent short-range SWM organization despite interindividual anatomical variability.

# 1 Introduction

The superficial white matter (SWM), situated between the deep white matter and the cortex, contains a dense network of short-range association fibers that connect within and between gyri [1–7]. Although the shape of the SWM connections varies [7–9], the most prominent U-shaped fibers (U-fibers) account for approximately twice the volume of long-range white matter bundles that have traditionally dominated diffusion MRI (dMRI) tractography research [10]. Recent dMRI tractography studies have underscored the SWM's importance for cognition, its development across the lifespan, and its vulnerability to aging and disease [11–18]. Despite growing interest in the SWM [19], most dMRI studies have focused on relatively deeper and longer bundles, which are more readily captured with standard clinical and research imaging protocols [3,20]. However, recent advances in high-resolution (submillimeter) dMRI have made it possible to resolve shorter SWM connections located closer to the cortical surface [8]. These short fibers are more variable across individuals due to differences in cortical folding, making their study both technically challenging and biologically informative. Their short trajectories and close relationship to highly variable cortical folding also make the identification of corresponding connections across individuals particularly challenging.

Several groups have addressed the challenge of characterizing population-level SWM organization by using fiber clustering methods to organize streamlines into bundles, demonstrating the feasibility of generating tractography-based population atlases of the SWM [3,9,21–25]. However, multiple challenges remain. Current SWM fiber clustering studies have prioritized longer bundles, generally ranging from 30 to 85 mm in length [9,21,24]. Although several SWM clustering methods exist, none to date have incorporated ultra-high-resolution submillimeter dMRI data, which significantly improves the reconstruction of shorter fibers [8]. Therefore, shorter bundles remain under-characterized, despite their abundance in the brain. These short-range connections also exhibit substantial geometric diversity, including not only canonical U-shaped fibers but also J-shaped, intra-gyral, and other local configurations [7,9,26], further complicating their population-level characterization.

Furthermore, unlike deep white matter tracts, which benefit from well-established anatomical nomenclature, SWM connections are generally defined by the cortical regions they connect [20]. Therefore, SWM fiber clustering methods typically group streamlines based on trajectory similarity, often guided by a cortical parcellation for cluster annotation or initialization. These parcellations can also aid in identifying corresponding clusters across subjects. However, most methods rely on a single, relatively coarse cortical atlas, such as the Desikan-Killiany atlas or its variants [3,9,21–23,25], which may lack the resolution and specificity needed to capture the full diversity of SWM trajectories, including intra-gyral connections and variations on the classic U-shape, such as J-shaped fibers [7,9]. More fundamentally, a single parcellation provides only one discretization of cortical anatomy; complementary parcellations may provide distinct and potentially useful descriptions of the cortical regions connected by a streamline. Moreover, cortical endpoint labels alone do not capture geometric differences among SWM trajectories, and fibers connecting the same cortical regions can exhibit diverse shapes. This motivates the incorporation of geometric characteristics of streamlines including their full trajectories and shape features [3,5]. Characterizing short-range SWM organization therefore requires the joint consideration of multiple sources of information, including fiber trajectory, streamline shape, and cortical anatomy represented by complementary parcellation schemes. Rather than viewing this solely as a fiber clustering problem, we frame it as a multimodal representation learning problem in which heterogeneous geometric and anatomical information is integrated to learn population-consistent representations of SWM connections.

Deep learning approaches have shown success in addressing hard unsupervised clustering problems, evolving from early two-stage pipelines [27,28] to joint representation-clustering frameworks [29,30]. More broadly, deep representation learning enables complex data to be mapped into latent feature spaces in which biologically or anatomically meaningful relationships can be more readily identified [31–35]. Building on this, multimodal deep learning extends these approaches to heterogeneous data (e.g., images, text, biomedical signals) to improve robustness [36]. LLMs have advanced rapidly in recent years, demonstrating a strong capacity for learning semantic representations from text and showing successful applications in the medical field [37,38]. More recently, multimodal representation learning, including vision-language models, has demonstrated the ability to learn unified representations across heterogeneous data types and to generalize across image-text tasks [39].

Previous studies have demonstrated successful applications of deep learning to unsupervised fiber clustering tasks [40–42]. For example, Deep Fiber Clustering (DFC) solves the unsupervised fiber clustering problem as a self-supervised learning task with a domain-specific pretext task to predict pairwise fiber distances and leverages both dMRI tractography and cortical parcellation information to enhance the anatomical coherence of the resulting fiber clusters [40]. However, several limitations exist in the DFC framework. First, DFC only studied streamlines longer than 40 mm, while the understudied shorter SWM bundles were ignored. Second, although DFC incorporates cortical parcellation information as an anatomical constraint during fiber clustering, its influence on the learned representation is limited by the method design. In addition, only one coarse cortical atlas is used in DFC, providing limited anatomical context for distinguishing the diverse short-range connections of the SWM. For the substantially shorter and more geometrically variable fibers revealed by submillimeter dMRI, richer anatomical information may be particularly important for establishing correspondence across individuals.

Advances in LLMs suggest a new opportunity for neuroanatomical modeling: information represented using different anatomical labeling schemes can potentially be expressed through language and embedded into a common continuous feature space rather than treated as unrelated categorical labels. Instead of encoding parcel identities from different cortical atlases as independent categorical variables, the anatomical context of each streamline can be described in text and mapped into a language-derived continuous representation. This representation can then be integrated with learned representations of streamline geometry and explicit shape features. To our knowledge, language-model representations of cortical anatomy have not previously been incorporated into tractography-based modeling of white matter organization. This approach enables information from multiple complementary cortical parcellations, which need not share the same nomenclature, spatial resolution, or parcel boundaries, to contribute jointly to a learned representation of each connection.

In this work, we present FiberGeoText (FGT), a language-informed multimodal deep embedding framework for learning population-level organization of short-range SWM connections. First, the proposed framework is designed for the important but understudied short SWM fibers, which are reconstructed from our acquired ultra-high-resolution (submillimeter) dMRI data. Second, FiberGeoText learns a joint embedding of three complementary aspects of each streamline: input dMRI tractography describing its three-dimensional trajectory, anatomical cortical parcellations describing its cortical context, and streamline shape features describing its local morphology. Third, anatomical information of endpoint regions from multiple cortical parcellations is expressed as text and encoded using a pretrained language model, providing a common language-based representation of heterogeneous anatomical descriptions. These anatomical representations are integrated with learned fiber geometry and explicit shape features in a shared embedding space in which population-consistent SWM connections can be identified across subjects. Thus, rather than defining SWM organization from fiber geometry or a single cortical parcellation alone, FiberGeoText learns a multimodal embedding of SWM correspondence from complementary geometric, anatomical, and shape information using a vision-language model. The learned embedding is subsequently partitioned into population-level clusters to provide consistent connection identities across subjects. Finally, the proposed framework achieves improved anatomical coherence, shape consistency, cross-subject consistency and geometric compactness. Evaluation in unseen testing subjects further supports its generalization across individuals.

# 2 Methods

We developed FiberGeoText to learn a multimodal representation of short-range SWM streamlines from complementary geometric, anatomical, and shape information. Figure 1 provides an overview of the framework. For each streamline, we characterize three aspects of the connection: (1) its three-dimensional trajectory derived from high-resolution dMRI tractography, (2) its cortical anatomical context in the form of text derived from multiple cortical parcellations, and (3) its shape using streamline-level geometric features. Fiber trajectory and cortical anatomy are encoded using a point-based neural network and an LLM respectively, and combined with the shape features to form a joint streamline embedding. The learned embeddings are then used to derive a population-level organization of SWM connections, with discrete clusters providing consistent connection identities that can subsequently be assigned to streamlines in new subjects. Below, we first describe the imaging data and multimodal streamline information (Section 2.1), followed by the FiberGeoText multimodal deep embedding framework and population-level organization strategy (Section 2.2).

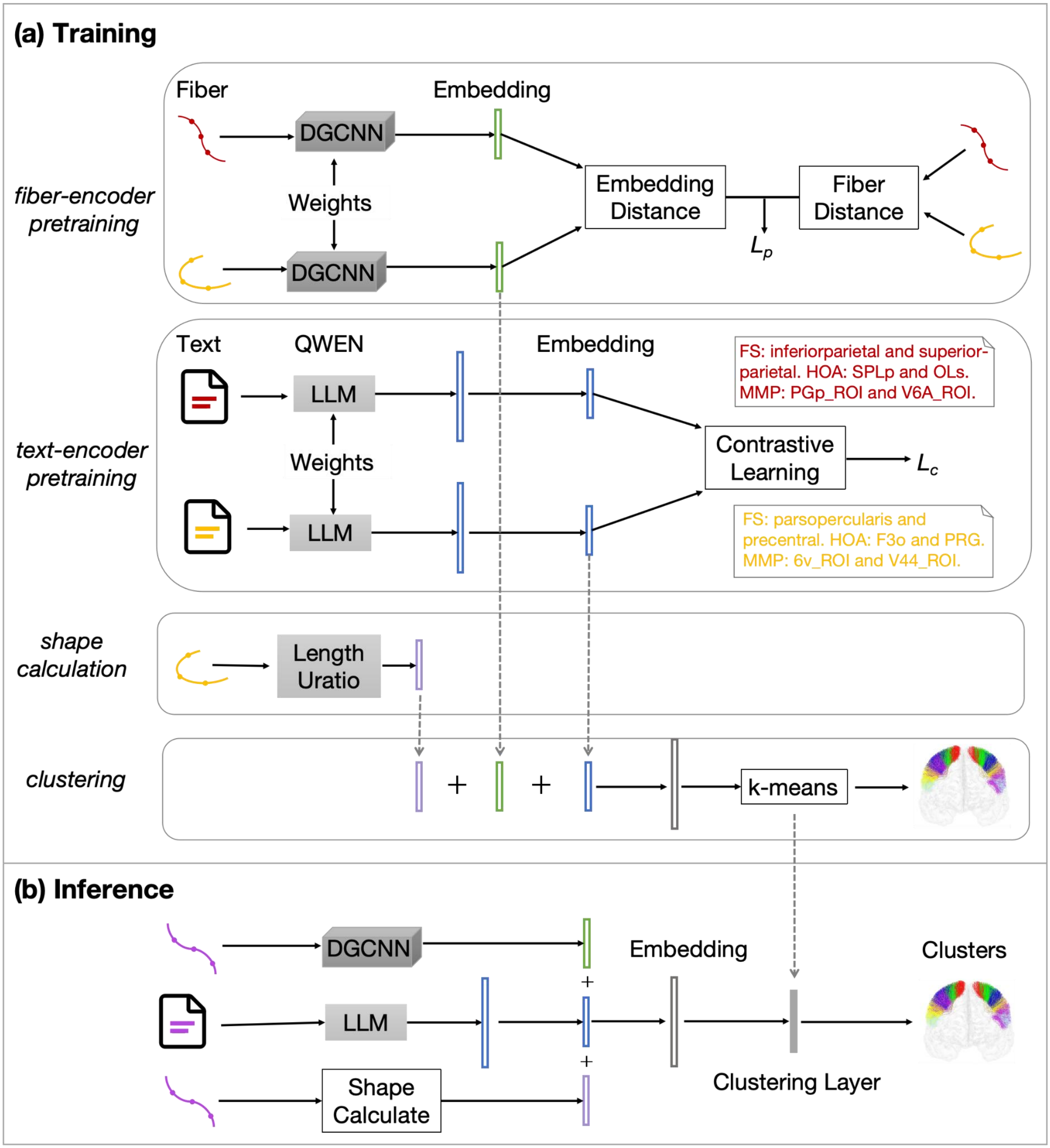

Figure 1 Overall pipeline.

## 2.1 Imaging data and multimodal streamline information

### 2.1.1 High-resolution dMRI

We evaluated our FGT method on a unique, ultra-high-resolution submillimeter diffusion MRI (dMRI) dataset comprising scans from 10 healthy adult participants (mean age 36 ± 7.5 years; 5 male and 5 female), collected at Brigham and Women's Hospital in Boston, USA. The study protocol was approved by the Mass General Brigham Institutional Review Board, and informed consent was obtained from all participants prior to imaging. Each subject underwent dMRI acquisition on a Siemens Prisma 3T scanner using the gSlider sequence (TE = 87 ms, TR = 3500 ms, voxel size = 0.76 × 0.76 × 3.8 mm³, in-plane acceleration factor = 3, multi-band factor = 2, and 6/8 partial Fourier encoding) [43,44]. Diffusion encoding was performed along 60 directions at a b-value of 1000 s/mm², along with three non-diffusion-weighted (b = 0) volumes. The standard gSlider reconstruction method was applied to obtain data with isotropic 0.76 mm³ resolution. High-resolution structural T1-weighted images were also collected (voxel size = 0.7 × 0.7 × 0.7 mm³, TE = 2.3 ms, and TR = 2500 ms).

For processing the gSlider dMRI data, we employed our established public diffusion MRI pipeline, available at https://github.com/pnlbwh/pnlutil. We applied FSL's topup and eddy tools [45] to correct for eddy currents, head motion, and EPI-induced distortions. Additional preprocessing steps included brain masking of T1-weighted (T1w) images using HD-BET [46] and dMRI brain masking using a CNN-based method [47].

### 2.1.2 SWM tractography

We employed *parallel transport tractography* (PTT) implemented in the trekker package[1], which provides smooth probabilistic tractography with high accuracy [48–50]. Parameters were tuned for our high-resolution data: stepSize=0.012, probeLength=0.21, minRadiusOfCurvature=0.42, minDataSupport=0.08, and probeRadius=0.42. Streamlines of 6–50 mm were retained. Streamlines were seeded uniformly throughout the white matter and propagated bidirectionally until entering the cortex. Tracking was terminated after entering the cortex. To restrict tracking to association streamlines, only streamlines with both endpoints in cortex and confined to a single hemisphere were retained [51].

Streamlines were registered to MNI space. First, the b0 image of each subject was registered to the MNI [52] T2-weighted template ( ICBM 152 2009b Nonlinear Symmetric atlas ) using nonlinear registration with ANTs [53]. The resulting transforms were then applied to the streamlines in subject space to obtain tractography in MNI space. Eight subjects were used for training, with 150,000 streamlines sampled from each training subject, and two subjects were held out for testing, with 1,000,000 streamlines sampled from each testing subject.

### 2.1.3 Cortical anatomical information

We parcellatd the cerebral cortex on the T1w volumes using three atlases: the Desikan-Killiany (DK) atlas [54], the Harvard-Oxford Atlas 2.0 (HOA2.0) [55], and the Human Connectome Project (HCP) multi-modal parcellation (MMP) atlas [56]. The atlases differ in the number of parcels per hemisphere: the DK atlas identifies 34 regions, the HOA2.0 separates 73 regions, the Julich Brain atlas has 146 areas (including gap regions), and the MMP divides the cortex into 180 regions. For the HOA2.0, we trained a FreeSurfer classifier on five hand-parcellated brains; the other parcellations are publicly available. All three atlases were applied through FreeSurfer on the reconstructed cortical surfaces, converted to voxel-wise labelmaps and registered to dMRI space.

---

[1] https://dmritrekker.github.iopubliclyHuman Connectome Project/

To describe the cortical anatomy associated with each streamline, we recorded, for each endpoint, the anatomical region label intersected in each cortical atlas.

### 2.1.4 Streamline shape information

We quantified two complementary streamline shape features: length and U-ratio. Streamline length (mm), commonly used to select association fibers [3,21,51], was computed along the full streamline trajectory. The U-ratio, which has been used to characterize superficial streamline shape [57,58], was defined as the Euclidean distance between streamline endpoints divided by streamline length, with lower values indicating more curved, U-shaped streamlines and values near 1 indicating straighter trajectories.

## 2.2 FiberGeoText multimodal deep embedding framework

We propose FiberGeoText, a multimodal embedding framework that learns population-consistent representations of SWM connections by integrating three complementary sources of information for each streamline: its three-dimensional trajectory, cortical anatomical context, and shape. The overall pipeline is shown in Figure 1. FiberGeoText first learns a multimodal representation of each streamline, which is subsequently used to derive a population-level organization of SWM connections. Streamline geometry and cortical anatomy are encoded independently using two specialized neural networks: a point-based network for the streamline trajectory and a large language model for anatomical information derived from multiple cortical parcellations. These networks produce a geometry embedding and an anatomy embedding for each streamline, respectively. The two learned embeddings are then combined with the streamline shape information described in Section 2.1.4 to form a joint multimodal representation. After learning the joint embedding, k-means clustering [59] is performed in this joint embedding space to identify population-level SWM clusters. The resulting cluster centroids are incorporated into a clustering layer, enabling streamlines from new subjects to be assigned directly to the learned clusters without rerunning k-means [40].

### 2.2.1 Streamline geometry encoding

The first component of FiberGeoText encodes the three-dimensional geometry of each streamline. White matter fiber geometry is the most widely used information for fiber clustering. In this study, streamline geometry is represented by the sequence of three-dimensional point coordinates along each trajectory. Each input streamline is represented as a point cloud and encoded using a Dynamic Graph Convolutional Neural Network (DGCNN) to obtain a geometry embedding (Figure 1). To achieve bilateral clustering by identifying homologous structures in both hemispheres, we take the absolute values of the left-right coordinates of all points before encoding each streamline [40]. The geometry encoder is learned in a self-supervised manner using a domain-specific pretext task that predicts the geometric distance between pairs of input streamlines. In this way, streamlines with similar trajectories are encouraged to occupy nearby locations in the learned geometry embedding space. Further details of the architecture and self-supervised training strategy are provided in our previous work [40,42].

### 2.2.2 Cortical anatomy encoding using an LLM

To provide complementary anatomical context for the learned streamline representation, we incorporate the cortical anatomical information described in Section 2.1.3. For each streamline, the endpoint labels obtained from all three cortical parcellations are converted into a textual description of the cortical regions connected by the streamline (Figure 1). This provides a common language-based format for combining anatomical information from parcellation schemes with different region names and definitions. To enable bilateral clustering, we remove hemisphere information from the cortical region names in the textual description.

The resulting text is encoded using an LLM, the pretrained Qwen-0.6B embedding model [60], to obtain an initial 1024-dimensional language embedding. Qwen-0.6B embedding model is a general-purpose LLM with 0.6 billion parameters designed to generate semantically informative representations of textual inputs [60]. To adapt the LLMmodel for encoding SWM cortical anatomy, we introduce a trainable linear projection layer and fine-tune the model using an unsupervised contrastive learning approach. The adapted model maps the language embedding to an anatomy embedding with the same dimensionality as the geometry embedding. To construct training pairs based on the anatomical characteristics of our data, we propose a novel definition of positive and negative pairs of input streamlines. Specifically, a pair of streamlines is considered a positive sample if the streamlines connect the same unordered pair of cortical regions in at least one of the three parcellations; all other pairs are considered negative samples. To prevent the much larger number of anatomically dissimilar streamline pairs from dominating training, equal numbers of positive and negative pairs are sampled. The resulting anatomy embedding therefore captures similarities in cortical connectivity across multiple complementary parcellation schemes and is subsequently integrated with the geometry embedding.

### 2.2.3 Multimodal integration of geometry, anatomy, and shape

The geometry and anatomy embeddings capture complementary aspects of each SWM connection: the geometry embedding describes the spatial trajectory of the streamline, whereas the anatomy embedding describes the cortical regions associated with its endpoints. We additionally incorporate streamline length and U-ratio, defined in Section 2.1.4, to explicitly represent aspects of streamline shape that are particularly relevant to the heterogeneous morphology of short SWM fibers.

For each streamline, the geometry embedding, anatomy embedding, and two shape measures are concatenated to form the joint multimodal representation of the SWM connection. This representation is subsequently used to derive the population-level cluster organization. Before concatenation, the geometry and anatomy embeddings and shape measures were normalized to the range [0, 1] to control their relative contributions to distance in the joint embedding space. Thus, the organization of the resulting representation space is informed simultaneously by trajectory similarity, cortical anatomical similarity across multiple parcellations, and streamline shape.

### 2.2.4 Population-level organization and inference

After obtaining the joint multimodal embedding, we perform k-means clustering on the training streamlines. We evaluated the number of clusters k as described in Section 3.2.1 and selected k=5,000 for the final model and subsequent experiments. The resulting cluster centroids provide discrete connection identities within the population-level SWM organization learned from the training subjects. These centroids are stored in a clustering layer for subsequent application to unseen subjects.

During inference, the geometry and anatomy embeddings are generated for each streamline using the pretrained encoders and combined with its length and U-ratio to produce the same multimodal representation used during training. Each streamline is then assigned to the cluster whose centroid is closest in the learned embedding space. This avoids rerunning k-means clustering for each new subject and preserves the identities of the learned clusters across subjects, enabling population-consistent SWM correspondence [40].

# 3 Experiments and Results

## 3.1 Evaluation Metrics

We adopted multiple metrics to quantitatively evaluate the population-level SWM organization produced by the proposed framework. These metrics enable evaluation of the quality of a clustering solution from several perspectives, including anatomical coherence, cluster-size consistency, shape coherence, cluster compactness, cross-subject consistency, and generalization ability to unseen subjects.

**Cortical Parcel Coherence (CPC)**. We propose a new metric, CPC, to evaluate the anatomical coherence of streamline terminations within a cluster. For each cluster, CPC is defined as the proportion of streamlines terminating in the most frequently occurring unordered pair of cortical parcels. Higher CPC indicates greater consistency of cortical terminations within the cluster. The overall CPC is computed as the mean across clusters and was calculated separately for each cortical parcellation (HOA, DK, and MMP).

**Coefficient of Variation (CoV)** [61] **of Number of Streamlines (NoS).** Consistency in cluster size provides a complementary measure of the organization of the clustering solution. The coefficient of variation measures variability relative to the mean and is defined as the standard deviation divided by the mean. We calculated the CoV of NoS across clusters as the standard deviation of cluster NoS divided by the mean cluster NoS. A lower CoV of NoS indicates more uniform cluster sizes.

**Coefficient of Variation (CoV) of U-ratio and length.** We expect streamlines within a cluster to have similar geometric characteristics, including U-ratio and length. Therefore, we calculate the CoV of U-ratio and length within each cluster. The overall CoV for each measure is calculated as the mean CoV across all clusters. Lower values indicate greater within-cluster shape consistency.

**Percentage of Subjects (PoS) and Percentage of Clusters (PoC).** To evaluate cross-subject consistency of the learned clusters in the training data, we calculate the Percentage of Subjects (PoS) represented within each cluster. PoS is defined as the proportion of training subjects contributing at least one streamline to a given cluster, and the overall PoS is calculated as the mean across all clusters. Higher PoS indicates that clusters are more consistently represented across the training population. For unseen testing subjects, we use the complementary Percentage of Clusters (PoC) metric to evaluate generalization. PoC is defined as the proportion of learned clusters that are detected in each testing subject. Higher PoC indicates that a larger proportion of the population-level cluster organization can be recovered in unseen individuals.

**Root Normalized Within-Cluster Sum of Squares (RNWCSS).** Within-Cluster Sum of Squares (WCSS) [62] is a widely used measure of cluster compactness [63]. To evaluate geometric compactness in streamline space, RNWCSS is calculated as the square root of the mean within-cluster squared streamline distance, as defined below:

$$\mathrm{WCSS} = \sum_{i=1}^{N} \mathrm{d}\,(f_i, c_i)^2, \ \mathrm{RNWCSS} = \sqrt{\frac{\mathrm{WCSS}}{N}}$$

where *N* is the total number of fibers, $f_i$ is the i-th fiber, $c_i$ is the centroid fiber of the corresponding cluster, and $\mathrm{d}\,(\cdot,\cdot)$ is the calculation of fiber distance (minimum average direct-flip distance is adopted here [64]). RNWCSS quantifies within-cluster geometric dispersion. Lower values indicate that streamlines are closer to their assigned cluster centroid, corresponding to greater geometric compactness.

**Distortion ratio** [65]**.** To further evaluate generalization to unseen subjects, we calculate a distortion ratio that measures how well testing-subject embeddings conform to the cluster structure learned from the training data. Specifically, the distortion ratio is the ratio of the mean squared embedding-to-centroid distance in the testing data to that in the training data, using the cluster centroids learned from the training set in both cases. A value near 1 indicates comparable embedding-to-centroid distances in training and testing data, whereas larger values indicate greater mismatch between unseen-subject embeddings and the learned cluster organization.

## 3.2 Experimental Results

We performed four experimental evaluations to characterize different aspects of the proposed framework: (1) evaluation of the effect of the number of clusters, (2) comparison with state-of-the-art methods, (3) ablation studies, and (4) generalization to unseen testing subjects. We leveraged the unique ultra-high-resolution dMRI dataset of 10 subjects acquired at 0.76-mm isotropic resolution, using eight subjects for training and two entirely held-out unseen subjects for testing. We utilized 1,200,000 streamlines across 8 subjects for model training and population-level clustering and applied the resulting model and learned cluster organization to the remaining unseen subjects (1,000,000 streamlines each). Experiments 1–3 were performed using the training data, while Experiment 4 evaluated generalization using the held-out testing subjects.

### 3.2.1 Effect of the Number of Clusters

We performed experiments to evaluate the effect of clustering granularity on the proposed method, as shown in Table 1. We compared values of k ranging from 500 to 10,000 (k = 500, 1,000, 1,500, 2,500, 5,000, 7,500, and 10,000), spanning and extending beyond the numbers of SWM bundles or clusters reported in previous studies [9,24]. As the number of clusters increases from 500 to 10,000, CPC consistently improves across all three cortical parcellation schemes, indicating progressively greater anatomical coherence within increasingly fine-grained clusters. At the same time, the CoVs of U-ratio and streamline length decrease, indicating greater within-cluster shape consistency. RNWCSS also decreases, reflecting greater geometric compactness. These improvements are accompanied by tradeoffs: the CoV of NoS increases, indicating greater variability in cluster size, while PoS decreases from 1.0 to 0.857, indicating that finer clusters are less consistently represented across training subjects. Based on this tradeoff between within-cluster coherence and cross-subject consistency, we selected k = 5,000 for subsequent experiments.

### 3.2.2 Comparison with SOTA Fiber Clustering Methods

We compared FiberGeoText (FGT) with two existing fiber clustering methods, DFC and FFClust. DFC is our previous deep fiber clustering framework, upon which FGT builds [40,42]. DFC incorporates cortical parcel similarity as an anatomical constraint during clustering, while its learned streamline representation is derived from tractography geometry. In contrast, FGT learns a multimodal streamline representation that explicitly integrates cortical endpoint anatomy, learned fiber geometry, and shape information before population-level clustering. FFClust [66] is an unsupervised fiber clustering method that has recently been applied to construct an SWM atlas [24]. For FGT and DFC, we generated 5,000 clusters, using the value of k selected in Section 3.2.1. For FFClust, we adjusted its parameters to obtain a comparable number of clusters, resulting in 4,918 clusters.

Table 1  Comparison of results among different numbers of clusters.

| number of clusters | | k=500 | k=1000 | k=1500 | k=2500 | k=5000 | k=7500 | k=10000 |
|---|---|---|---|---|---|---|---|---|
| CPC | DK ↑ | 0.849 | 0.898 | 0.915 | 0.935 | 0.955 | 0.965 | 0.971 |
| | HOA ↑ | 0.639 | 0.728 | 0.767 | 0.822 | 0.879 | 0.904 | 0.919 |
| | MMP ↑ | 0.252 | 0.298 | 0.330 | 0.370 | 0.437 | 0.484 | 0.516 |
| CoV of NoS ↓ | | 0.354 | 0.394 | 0.419 | 0.442 | 0.485 | 0.502 | 0.526 |
| CV of U-ratio ↓ | | 0.174 | 0.161 | 0.154 | 0.144 | 0.133 | 0.124 | 0.119 |
| CV of length ↓ | | 0.243 | 0.225 | 0.211 | 0.198 | 0.180 | 0.170 | 0.163 |
| PoS ↑ | | 1 | 0.99975 | 0.995 | 0.9823 | 0.9385 | 0.897 | 0.857 |
| RNWCSS ↓ | | 10.425 | 9.432 | 8.906 | 8.241 | 7.425 | 6.986 | 6.671 |

The results of SOTA comparison are shown in Table 2. FGT achieves the highest anatomical coherence across all three cortical parcellations, as indicated by the highest CPC values for DK, HOA, and MMP. FGT also shows lower variability in cluster size and substantially greater within-cluster shape consistency, reflected by lower CoV values for NoS, U-ratio, and streamline length than DFC and FFClust. PoS is also highest for FGT, indicating that its clusters are represented across a larger proportion of training subjects on average. DFC achieves the lowest RNWCSS, indicating greater purely geometric compactness, whereas FGT exhibits higher RNWCSS. This reflects an expected tradeoff of the multimodal formulation: FGT allows greater spatial geometric variability within a cluster in order to identify anatomically corresponding streamlines across subjects despite interindividual anatomical variability. Rather than optimizing geometric proximity alone, FGT jointly considers cortical anatomy, fiber geometry, and streamline shape, enabling anatomically coherent correspondences even when their spatial trajectories do not align exactly across individuals.

Figure 2 visualizes example clusters from training data for FGT, DFC, and FFClust. For visualization, corresponding clusters in DFC and FFClust were identified as the geometrically closest clusters to each selected FGT cluster. The examples illustrate greater within-cluster shape similarity and endpoint convergence for FGT, consistent with the quantitative results.

Table 2 Comparison of population-level SWM organization across FiberGeoText and state-of-the-art fiber clustering methods (↑ higher is better; ↓ lower is better).

| methods | | FGT (k=5000) | DFC (k=5000) | FFClust (k=4918) |
|---|---|---|---|---|
| CPC | DK ↑ | **0.955** | 0.786 | 0.530 |
| | HOA ↑ | **0.879** | 0.602 | 0.672 |
| | MMP ↑ | **0.437** | 0.403 | 0.358 |
| CoV of NoS ↓ | | **0.485** | 0.854 | 2.158 |
| CoV of U-ratio ↓ | | **0.133** | 0.186 | 0.195 |
| CoV of length ↓ | | **0.180** | 0.263 | 0.190 |
| PoS ↑ | | **0.9385** | 0.9022 | 0.699 |
| RNWCSS ↓ | | 7.425 | **4.716** | 6.395 |

### 3.2.3 Ablation study

We performed an ablation study to investigate the complementary contributions of fiber geometry, cortical anatomy, and streamline shape to the learned SWM organization. Four configurations were evaluated: Fiber-only, Text-only, Fiber+Text, and the full FGT framework. Fiber-only uses only the streamline geometry embedding derived from the WM geometry encoding subnetwork. Text-only uses only the language-derived cortical anatomy embedding. Fiber+Text concatenates the geometry and anatomy embeddings, while the full FGT additionally incorporates streamline length and U-ratio.

As shown in Table 3, Text-only, Fiber+Text, and FGT achieve substantially higher CPC values than Fiber-only across all three cortical parcellations. This confirms that cortical anatomical information strongly influences the organization of streamlines in the learned representation. However, Text-only produces considerably greater variability in cluster size and streamline shape and the highest RNWCSS, indicating that anatomical endpoint information alone does not sufficiently constrain streamline geometry or shape. Compared with Fiber+Text, adding explicit shape information in FGT substantially reduces the CoVs of U-ratio and length and increases PoS, while moderately reducing CPC. This demonstrates a tradeoff between maximizing endpoint-label coherence and obtaining clusters that are simultaneously coherent in anatomy, trajectory, and shape. To better evaluate the overall performance of these methods, we calculated an additional metric, Mean Normalized Score. This score was computed as the unweighted mean of all metrics after independently applying min–max normalization to [0,1]. For metrics where lower values indicate better performance, the normalization was reversed so that 1 and 0 consistently represent the best and worst performance, respectively. As shown in the table, FGT achieved the best overall performance, indicated by the highest Mean Normalized Score, followed by Fiber+Text. Overall, the ablation results demonstrate complementary roles for the three information sources: cortical anatomy drives endpoint coherence, fiber geometry preserves trajectory compactness, and explicit shape information increases within-cluster morphological consistency. Their integration in FGT produces the most balanced multimodal characterization of population-level SWM organization.

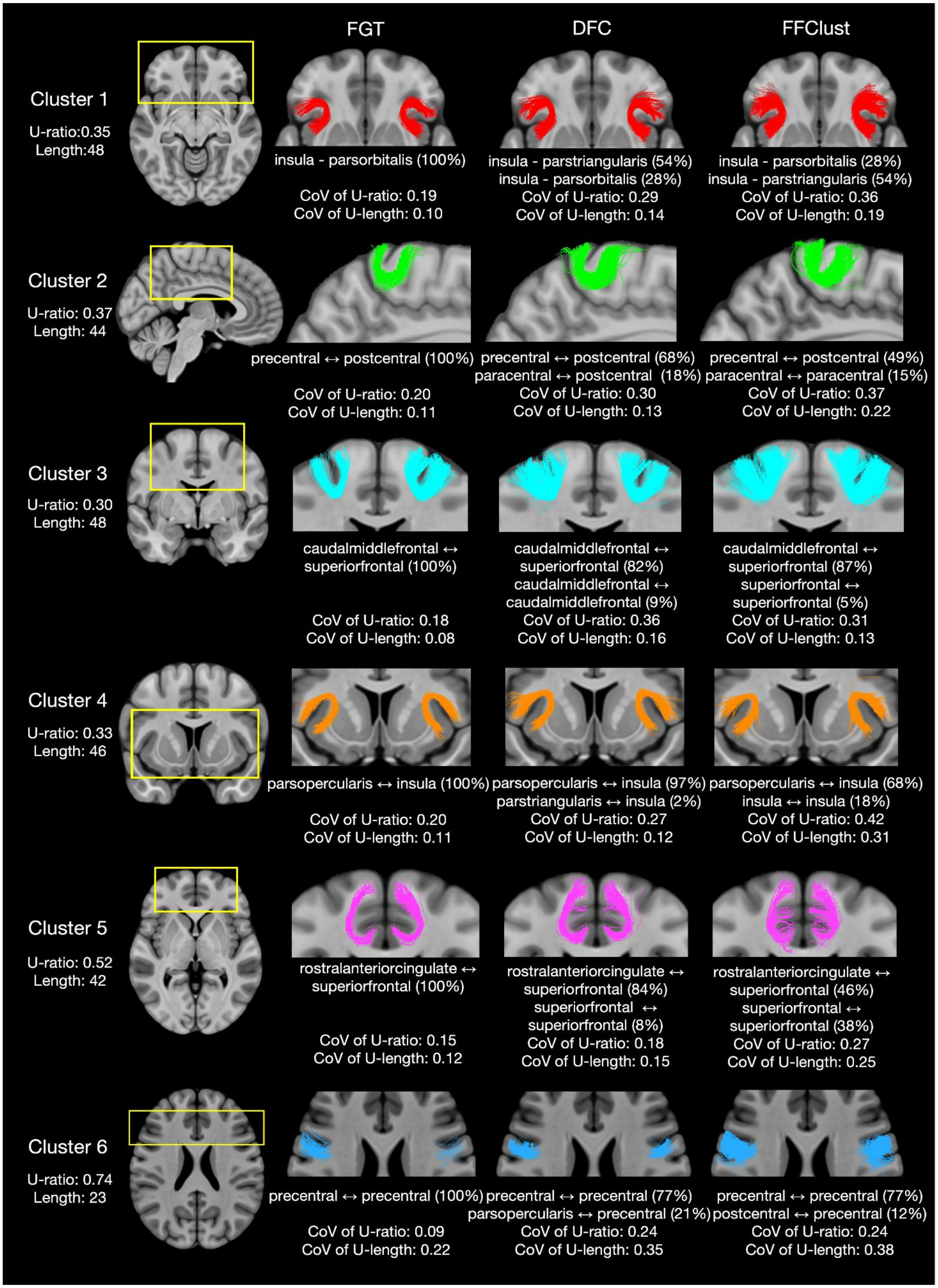

Figure 2. Visualization of example learned clusters from three methods (FGT, DFC, and FFClust), showing streamlines from training subjects. For clusters 1, 5, and 6, a superior view is displayed; for cluster 2, a right view is displayed; for clusters 3 and 4, an anterior view is displayed. For each example cluster, the U-ratio and fiber length, averaged across all methods, are annotated below the cluster number. Below each cluster visualization, the top two connected DK region pairs with the highest numbers of streamlines and their corresponding streamline percentages are reported, together with the CoV of U-ratio and length of the cluster.

Table 3 Ablation study of FiberGeoText (FGT). Arrows indicate the preferred direction of each metric (↑ higher is better; ↓ lower is better).

| methods | | Fiber-only | Text-only | Fiber+Text | FGT |
|---|---|---|---|---|---|
| CPC | DK ↑ | 0.716 | 0.964 | **0.970** | 0.955 |
| | HOA ↑ | 0.554 | **0.94** | 0.920 | 0.879 |
| | MMP ↑ | 0.362 | **0.631** | 0.511 | 0.437 |
| CoV of NoS ↓ | | **0.307** | 1.426 | 0.537 | 0.485 |
| CoV of U-ratio ↓ | | 0.202 | 0.227 | 0.212 | **0.133** |
| CoV of length ↓ | | 0.257 | 0.347 | 0.275 | **0.180** |
| PoS ↑ | | **0.9590** | 0.868 | 0.91125 | 0.9385 |
| RNWCSS↓ | | **4.174** | 10.249 | 7.060 | 7.425 |
| **Mean Normalized Score ↑** | | 0.372 | 0.476 | 0.611 | **0.768** |

## 3.2.4 Generalization to Unseen Subjects

We applied the trained FGT framework and learned population-level cluster centroids to the unseen testing subjects without reclustering. Generalization was evaluated by measuring the percentage of learned clusters detected in each testing subject (PoC) and by comparing the distribution of testing embeddings relative to the learned centroids using the distortion ratio.

Table 4 presents the quantitative evaluation results on the training data and two unseen testing subjects (sub-01 and sub-02). Overall, performance on the unseen subjects is comparable to that observed in the training data, indicating that the learned multimodal organization transfers well to new individuals. CPC values based on DK and HOA remain consistently high in both testing subjects. MMP-based CPC is higher in the individual testing subjects than in the pooled training data, which may reflect reduced anatomical variability when coherence is measured within a single subject. PoC is high in both unseen subjects, with 94.88% and 98.36% of the learned clusters detected in sub-01 and sub-02, respectively, indicating that most population-level clusters can be recovered in individual unseen subjects. The RNWCSS scores are also comparable between the training and testing data, indicating similar cluster compactness. Finally, the distortion ratios are close to 1 for both testing subjects, indicating that the unseen-subject embeddings have similar distances to the learned cluster centroids as the training embeddings.

Example clusters from training data and their corresponding clusters identified in the testing subjects are shown in Figure 3. The learned training clusters contain streamlines from multiple individuals, with each subject shown in a different color. Despite interindividual differences in streamline location and morphology, FGT identifies groups of trajectories with putatively corresponding anatomy and geometry across subjects. Corresponding clusters are also recovered in the two unseen testing subjects (right columns), providing qualitative evidence that the learned population-level organization generalizes across individuals.

Table 4 Clustering performance on unseen testing subjects.

| subjects | | training data | sub-01 | sub-02 |
|---|---|---|---|---|
| TSPC | DK | 0.955 | 0.954 | 0.951 |
| | HOA | 0.879 | 0.879 | 0.874 |
| | MMP | 0.437 | 0.572 | 0.523 |
| CoV of NoS | | 0.485 | 1.376 | 1.179 |
| CoV of U-ratio | | 0.133 | 0.119 | 0.120 |
| CoV of length | | 0.180 | 0.155 | 0.161 |
| PoC | | 0.9385 | 0.9488 | 0.9836 |
| RNWCSS | | 7.425 | 6.626 | 6.401 |
| Distortion ratio | | - | 1.2921 | 1.0775 |

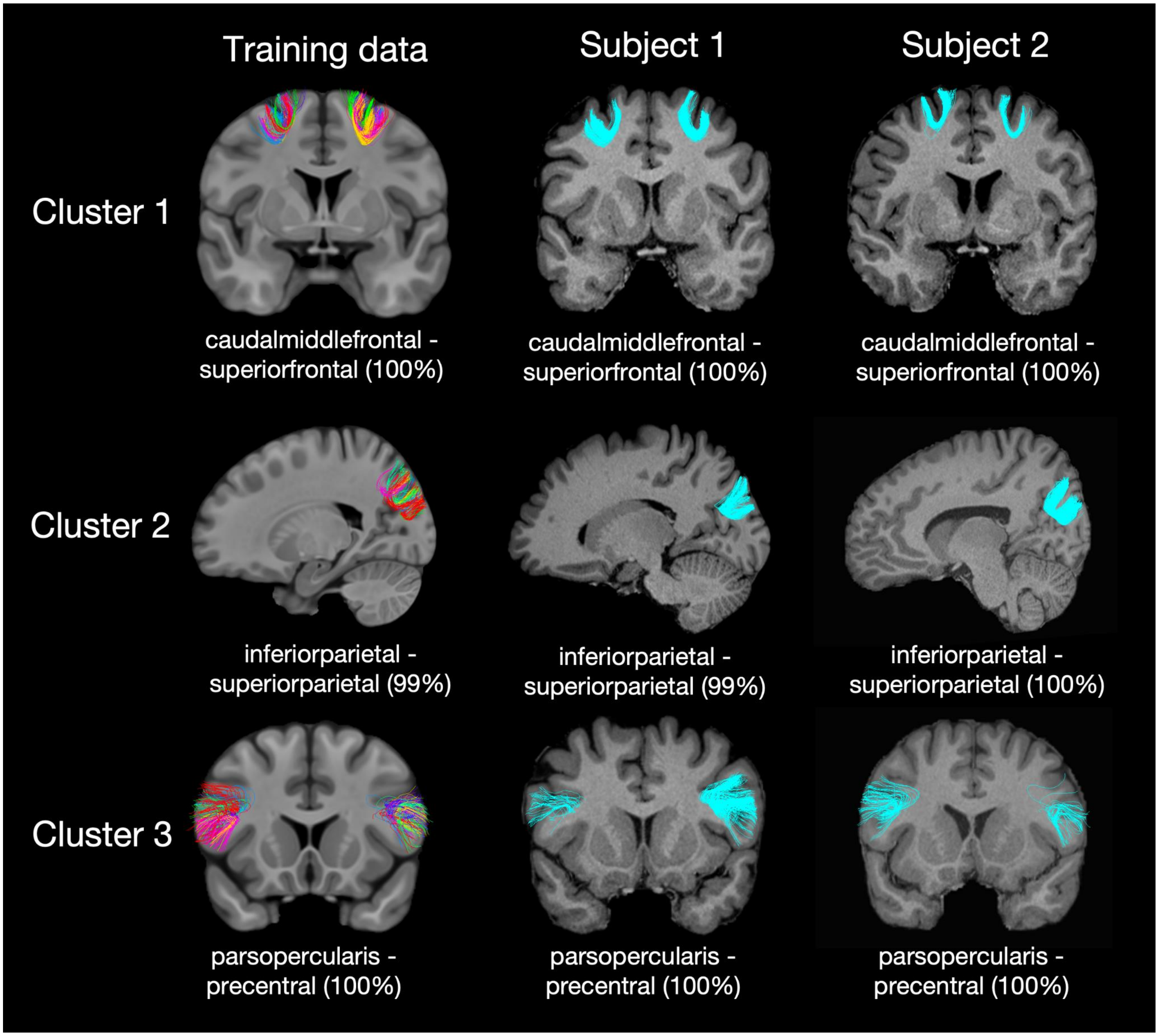

Figure 3. Visualization of example clusters obtained from training data and their corresponding clusters identified in the two testing subjects. The different colors of streamlines in the training data indicate different subjects, showing anatomical variability. Below each cluster visualization, the top two connected DK regions and their corresponding streamline percentages are reported.

# 4 Discussion

In this work, we introduced FiberGeoText (FGT), a multimodal deep embedding framework for population-level organization of short-range (6–50 mm) SWM streamlines reconstructed from submillimeter dMRI. FGT learns a joint representation that integrates streamline geometry, cortical anatomical information from multiple parcellations, and explicit shape measures. Our results show that these complementary sources of information enable short SWM streamlines to be organized across subjects into clusters that are coherent in cortical connectivity and morphology while reproducible across individuals.

An important finding is that SWM correspondence should not be defined by spatial proximity alone. DFC achieved lower RNWCSS, indicating greater purely geometric compactness, whereas FGT achieved substantially higher cortical parcel coherence and shape consistency. FGT therefore learns an embedding that permits greater intersubject spatial variation within a cluster in order to identify streamlines with corresponding cortical anatomy and morphology despite variability in cortical folding. This tradeoff is particularly relevant for the SWM, where homologous connections may not occupy precisely corresponding spatial locations across individuals. Rather than treating geometric compactness as the sole criterion for correspondence, FGT balances trajectory, cortical anatomy, and shape to obtain a population-level organization.

The cortical endpoint results further demonstrate the value of incorporating multiple anatomical descriptions. The learned clusters showed good cortical parcel coherence across parcellations of differing granularity, with CPC values of approximately 96%, 88%, and 44% for the Desikan–Killiany, Harvard–Oxford, and Glasser-MMP parcellations, respectively. The lower coherence for MMP likely reflects, at least in part, its substantially finer cortical subdivision, which provides a more stringent measure of endpoint agreement. Notably, MMP-based CPC was higher when evaluated within individual unseen subjects than in the pooled training data, suggesting that some of the lower population-level MMP coherence reflects intersubject variability in fine-scale cortical anatomy rather than poor within-subject anatomical organization. Importantly, FGT does not require the different parcellations to share a common nomenclature or set of region boundaries. By expressing endpoint information as text and embedding these descriptions with a pretrained LLM, multiple complementary anatomical labeling schemes can contribute to a common learned representation. Including multiple parcellations also enhances interpretability by allowing the learned clusters to be related to cortical anatomy across complementary labeling schemes and levels of granularity. In the present work, this includes the multimodally defined Glasser-MMP parcellation. The same strategy could, in the future, incorporate other anatomically relevant descriptions, including cytoarchitectonic parcellations, alternative cortical atlases, or parcellations selected for a particular neuroscientific or clinical question.

The ablation experiments clarify the complementary roles of the three information sources. Cortical anatomy strongly increased endpoint coherence but, when used alone, produced clusters with poorer geometric and shape consistency. Streamline geometry provided substantially greater spatial compactness, while adding length and U-ratio markedly improved within-cluster morphological consistency. Thus, no individual information source was sufficient to capture all aspects of SWM correspondence. The full FGT framework instead learned a representation jointly informed by where streamlines travel, which cortical regions they connect, and their local morphology. This result also suggests that explicit shape information is particularly useful for short SWM trajectories, whose morphology includes a continuum from relatively straight intra-gyral fibers to more strongly curved U-shaped configurations [9,26].

FGT also showed strong transfer of the learned population organization to unseen subjects. Approximately 96.7% of the 5,000 learned clusters on average were recovered in the testing subjects without reclustering, while distortion ratios remained close to those observed in the training data. Qualitative results similarly showed streamlines from different individuals grouping into putatively corresponding clusters despite differences in spatial location. Because inference consists of encoding a new streamline and assigning it to an existing cluster centroid, the learned organization can be applied directly to new subjects without repeating population-level clustering. Together, the multimodal representation learning and centroid-based inference in FGT enable large-scale organization of more than a million short-range streamlines in the present study and provide a practical basis for quantitative SWM analysis at spatial scales that have been difficult to study with previous clustering approaches.

Several limitations motivate future work. First, there is no direct ground truth for short-range SWM correspondence. The present evaluations quantify consistency of tractography geometry, shape, cortical anatomy, and cross-subject occurrence, but future work should investigate additional validation using anatomical dissection, tracer information where available, or functional and clinical associations. Second, FGT inherits uncertainties from both tractography reconstruction and cortical parcellation, and errors in either source may influence the learned organization. Third, the current framework concatenates geometry, anatomy, and shape information to form its multimodal representation before population-level clustering. Future work could investigate learned multimodal fusion or weighting strategies and incorporate additional anatomical information, including cytoarchitectonic or functionally relevant cortical parcellations. Finally, evaluation in larger and more diverse high-resolution datasets will also be important for characterizing population variability, refining the learned SWM organization, and establishing its generalizability across individuals.

# 5 Conclusion

In conclusion, our results show that short-range superficial white matter connections can be organized consistently across individuals by integrating complementary information about trajectory, cortical anatomy, and shape. We introduced FiberGeoText, a multimodal deep embedding framework for population-level organization of short-range SWM connections reconstructed from high resolution (submillimeter) diffusion MRI. By integrating learned fiber geometry, language-based representations of cortical anatomy, and explicit streamline shape features, FGT identifies population-consistent SWM organization that is coherent in cortical connectivity and morphology while accommodating interindividual spatial variability. These findings further indicate that SWM correspondence is not adequately captured by spatial proximity alone. The complementary contributions observed in our ablation experiments, together with the recovery of most learned clusters in unseen subjects, support a multimodal definition of SWM correspondence that is jointly based on trajectory, cortical anatomy, and shape. This framework can provide a new approach for population-scale quantitative characterization of short-range SWM organization and its future study in development, cognition, aging, and disease.